\documentclass[a4paper,twoside]{article}
\pdfoutput=1

\usepackage{epsfig}
\usepackage{subfigure}
\usepackage{calc}
\usepackage{amssymb}
\usepackage{amstext}
\usepackage{amsmath}
\usepackage{amsthm}
\usepackage{multicol}
\usepackage{pslatex}
\usepackage{apalike}

\usepackage{mathtools}
\usepackage[pagebackref=true,breaklinks=true,letterpaper=true,colorlinks,bookmarks=false]{hyperref}
\usepackage[capitalise,noabbrev]{cleveref}

\usepackage{SCITEPRESS}

\DeclareMathOperator*{\argmax}{argmax}
\DeclareMathOperator{\sign}{sign}
\DeclareMathOperator{\diag}{diag}

\newcommand{\eg}{\textit{e.g.}}
\newcommand{\ie}{\textit{i.e.}}

\begin{document}

    %%%%%%%%% TITLE
    \title{Automatic Query Image Disambiguation \\ for Content-Based Image Retrieval}
    
    \author{\authorname{Bj\"orn Barz and Joachim Denzler}
        \affiliation{Computer Vision Group, Friedrich Schiller University Jena, Ernst-Abbe-Platz 2, 07743 Jena, Germany \\ \em\url{http://www.inf-cv.uni-jena.de/}}
        \email{\{bjoern.barz, joachim.denzler\}@uni-jena.de}
    }
    
    
    %%%%%%%%% ABSTRACT
    \abstract{%
        Query images presented to content-based image retrieval systems often have various different interpretations, making it difficult to identify the search objective pursued by the user.
        We propose a technique for overcoming this ambiguity, while keeping the amount of required user interaction at a minimum. To achieve this, the neighborhood of the query image is divided into coherent clusters from which the user may choose the relevant ones. A novel feedback integration technique is then employed to re-rank the entire database with regard to both the user feedback and the original query.
        We evaluate our approach on the publicly available MIRFLICKR-25K dataset, where it leads to a relative improvement of average precision by 23\% over the baseline retrieval, which does not distinguish between different image senses.
    }
    
    \onecolumn \maketitle \normalsize \vfill

    %%%%%%%%% BODY TEXT
    \section{\uppercase{Introduction}}
    \label{sec:introduction}

    \noindent
    Content-Based Image Retrieval (CBIR) refers to the task of retrieving a ranked list of images from a potentially large database that are semantically similar to one or multiple given query images. It has been a popular field of research since 1993 \cite{niblack1993qbic} and its advantages over traditional image retrieval based on textual queries are manifold: CBIR allows for a more direct and more fine-grained encoding of what is being searched for using example images and avoids the cost of textual annotation of all images in the database. Even in cases where such describing texts are naturally given (\eg, when searching for images on the web), the description may lack some aspects of the image that the annotator did not care about, but the user searching for that image does.
    
    In some applications, specifying a textual query for images may even be impossible. An example is biodiversity research, where the class of the object on the query image is unknown and to be determined using similar images retrieved from an annotated database \cite{freytag2015interactive}. Another example is flood risk assessment based on social media images \cite{poser2010volunteered}, where the user searches for images that allow for estimation of the severity of a flood and the expected damage. This search objective is too complex for being expressed in the form of keywords (\eg, ``images showing street scenes with cars and traffic-signs partially occluded by polluted water'') and, hence, has to rely on query-by-example approaches.
    
    \begin{figure}[tb]
        \includegraphics[width=\linewidth]{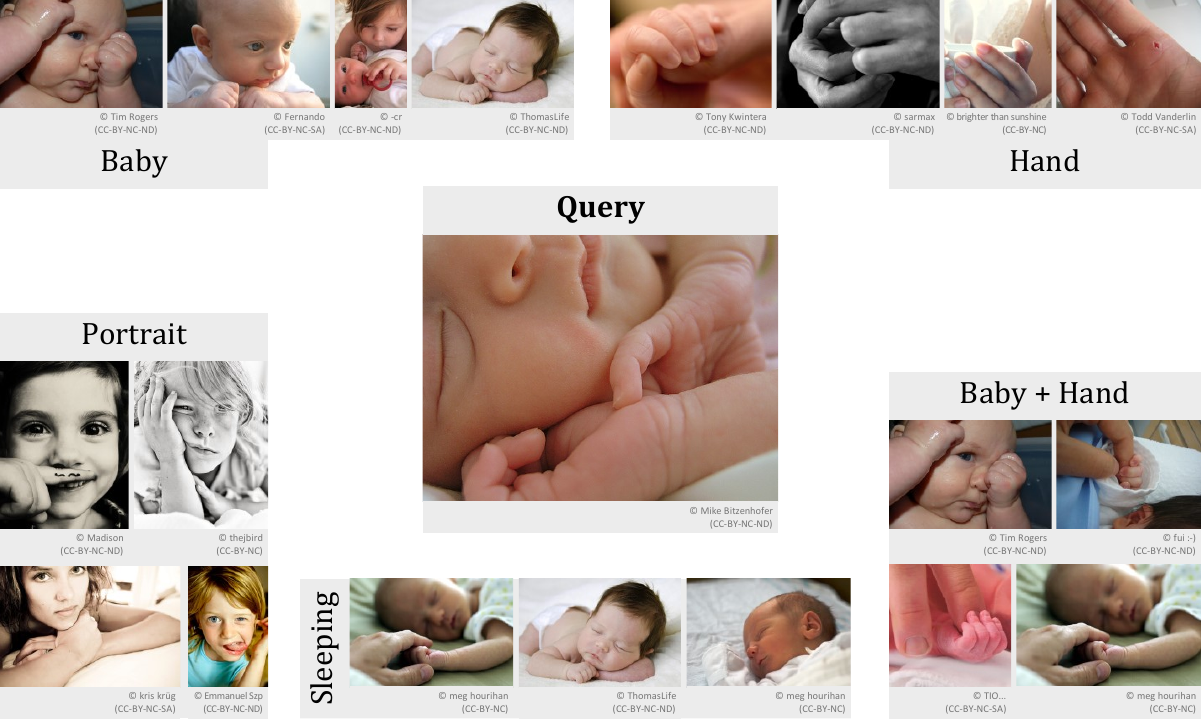}%
        \caption{An example of query image ambiguity. Given the query image in the center, the user might be searching for any of the topics listed around it, which do all appear intermixed in the baseline retrieval results. All images are from the MIRFLICKR-25K dataset \cite{huiskes08mirflickr}.}
        \label{fig:ambiguity-examples}
    \end{figure}
    
    In the recent past, there has been a notable amount of active research on the special case of \textit{object} or \textit{instance retrieval} \cite{jegou2010aggregating,arandjelovic2012three,jegou2014triangulation,babenko2015aggregating,yu2017exploiting,gordo2016end}, which refers to the task of only retrieving images showing exactly the same object as the query image. Approaches for solving this problem have reached a mature performance on the standard object retrieval benchmarks recently thanks to end-to-end learned deep representations \cite{gordo2016end}.
    
    However, in the majority of search scenarios, users are not looking for images of exactly the same object, but for images similar, but not identical to the given one. This involves some ambiguity inherent in the query on several levels (an example is given in \cref{fig:ambiguity-examples}):
    \begin{enumerate}
        \item Different users may refer to different regions in the image. This problem is evident if the image contains multiple objects, but the user may also be looking for a specific part of a single object.
        \item If the user is searching for images showing objects of the same class as the object in the query image, the granularity of the classification in the user's mind is not known to the system. If the query showed, for example, a poodle, the user may search for other poodles, dogs, or animals in general.
        \item A single object may even belong to multiple orthogonal classes. Given a query image showing, for instance, a white poodle puppy, it is unclear whether the user is searching for poodles, for puppies, for white animals, or for combinations of those categories.
        \item The visual aspect of the image that constitutes the search is not always obvious. Consider, for example, an oil painting of a city skyline at night as query image. The user may search for other images of cities, but she might also be interested in images taken at night or in oil paintings regardless of the actual content.
    \end{enumerate}
    
    \noindent
    Given all these kinds of ambiguity, it is often impossible for an image retrieval system to provide an accurate, satisfactory answer to a query consisting of a single image without any further information. Many CBIR systems hence enable the user to mark relevant and sometimes also irrelevant images among the initially retrieved results. This relevance feedback is then used to issue a refined query \cite{rocchio1971relevance,jin2003improving,deselaers2008learning}. This process, however, relies on the cooperation and the patience of the user, who may not be willing to go through a large set of mainly irrelevant results in order to provide extensive relevance annotations.
    
    In this work, we present an approach to simplify this feedback process and reduce the user's effort to a minimum, while still being able to improve the relevance of the retrieved images significantly. Our method consists of two steps: First, we automatically identify different meanings of the query image through clustering of the highest scoring retrieval results. The user may select one or more relevant clusters based on a few preview images shown for each cluster. We then apply a novel re-ranking technique that adjusts the scores of all images in the database with respect to this simple user feedback. Note that the number of clusters to choose from will be much smaller than the number of images the user would have to annotate for image-wise relevance feedback.
    
    Our re-ranking technique adjusts the effective distance of database images from the query, so that images in the same direction from the query as the selected cluster(s) are moved closer to the query and images in the opposite direction are shifted away. This avoids error-prone hard decisions for images from a single cluster and takes both the similarity to the selected cluster and the similarity to the query image into account.
    
    For all hyper-parameters of the algorithm, we propose either appropriate default values or suitable heuristics for determining them in an unsupervised manner, so that our method can be used without any need for hyper-parameter tuning in practice.
    %Our entire approach can be used without the need for any hyper-parameter tuning thanks to heuristics for automatic hyper-parameter determination and suitable default values, making the method effectively hyper-parameter free.
    
    The remainder of this paper is organized as follows: We briefly review related work on relevance feedback in image retrieval and similar clustering approaches in \cref{sec:related-work}. The details of our proposed \textit{Automatic Query Image Disambiguation (AID)} method are set out in \cref{sec:AID}. Experiments described in \cref{sec:experiments} and conducted on a publicly available dataset of 25,000 Flickr images \cite{huiskes08mirflickr} demonstrate the usefulness of our method and its advantages over previous approaches. \Cref{sec:conclusions} summarizes the results.

    \section{\uppercase{Related Work}}
    \label{sec:related-work}
    
    \noindent
    The incorporation of relevance feedback has been a popular method for refinement of search results in information retrieval for a long time. Typical approaches can be divided into a handful of classes:
    \textbf{Query-Point Movement (QPM)} approaches adjust the initial query feature vector by moving it towards the direction of selected relevant images and away from irrelevant ones \cite{rocchio1971relevance}. Doing so, however, they assume that all relevant images are located in a convex cluster in the feature space, which is rarely true \cite{jin2003improving}.
    On the other hand, approaches based on \textbf{distance or similarity learning} optimize the distance metric used to compare images, so that the images marked as relevant have a low pair-wise distance, while having a rather large distance to the images marked as irrelevant \cite{ishikawa1998mindreader,deselaers2008learning}. In the simplest case, the metric learning may consist in just re-weighting the individual features \cite{deselaers2008learning}.
    Speaking of machine learning approaches, \textbf{classification} techniques are also often employed to distinguish between relevant and irrelevant images in a binary classification setting \cite{guo2002learning,tong2001support}.
    Finally, \textbf{probabilistic} approaches estimate the distribution of a random variable indicating whether a certain image is relevant or not, conditioned by the user feedback \cite{cox2000bayesian,arevalillo2010naive,glowacka2016image}.
    
    However, all those approaches require the user to give relevance feedback regarding several images, which often has to be done repeatedly for successive refinement of retrieval results. Some methods even need more complex feedback than binary relevance annotations, asking the user to assign a relevance \textit{score} to each image \cite{kim2003qcluster} or to annotate particularly important regions in the images \cite{freytag2015interactive}.
    
    In contrast, our approach keeps the effort on the user's side as low as possible by restricting feedback to the selection of a single cluster of images.
    Resolving the ambiguity of the query by clustering its neighborhood has been successfully employed before, but very often relies on textual information \cite{zha2009visual,loeff2006discriminating,cai2004hierarchical}, which is not always available. One exception is the CLUE method \cite{chen2005clue}, which relies solely on image features and is most similar to our approach.
    In opposition to CLUE, which uses spectral clustering for being able to deal with non-metric similarity measures, we rely on k-Means clustering in Euclidean feature spaces, so that we can use the centroids of the selected clusters to refine the retrieval results.
    
    A major insufficiency of CLUE and other existing works is that they fail to provide a technique for incorporating user feedback regarding the set of clusters provided by the methods. Instead, the user has to browse all clusters individually, which is not optimal for several reasons: First, similar images near the cluster boundaries are likely to be mistakenly located in different clusters and, second, a too large number of clusters will result in the relevant images being split up across multiple clusters. Moreover, the overall set of results is always restricted to the initially retrieved neighborhood the query.
    
    Our approach is, in contrast, able to re-rank the entire dataset with regard to the selection of one or more clusters in a way that avoids hard decisions and takes both the distance to the initial query and the similarity to the selected cluster into account.

    \section{\uppercase{Automatic Query Image Disambiguation (AID)}}
    \label{sec:AID}
    
    \begin{figure*}[t]
        \includegraphics[width=\linewidth]{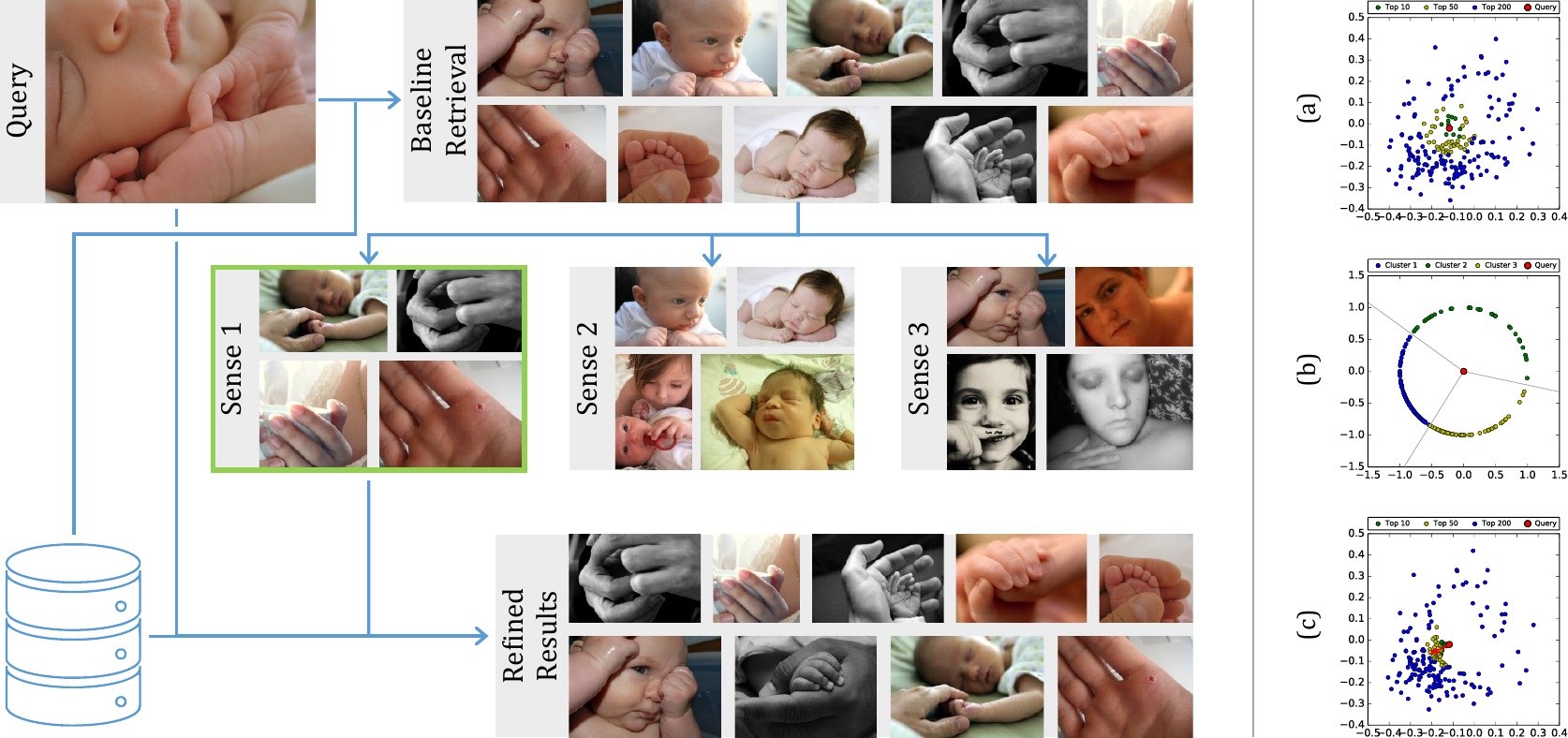}%
        \caption{\textbf{Left:} Illustration of query image disambiguation and refinement of retrieval results. In this example, the user selected Sense 1 as relevant. \textbf{Right:} Schematic illustration of the data representation at each step in 2-d space.}
        \label{fig:aid-example}
    \end{figure*}
    
    \noindent
    Our automatic query image disambiguation method (AID) consists of two parts: The unsupervised identification of different meanings inherent in the query image (cf.\ \cref{subsec:AID-clustering}), from which the user may then choose relevant ones, and the refinement of the retrieval results according to this feedback (cf.\ \cref{subsec:AID-refinement}). The entire process is illustrated exemplarily in \cref{fig:aid-example}.
    
    \subsection{Identification of Image Senses}
    \label{subsec:AID-clustering}
    
    In the following, we assume all images to be represented by $d$ real-valued features, which could be, for example, \textit{neural codes} extracted from a neural network (cf.\ \cref{subsec:setup}). Given a query image $q \in \mathbb{R}^d$ and a database $\mathcal{B} \subset \mathbb{R}^d$ with $n \coloneqq |B|$ images, we first retrieve the $m$ nearest neighbors $X = \{ x_1, x_2, \dots, x_m \} \subseteq \mathcal{B}$ of $q$ from the database. We employ the Euclidean distance for this purpose, which has been shown to be a reasonable dissimilarity measure when used in combination with semantically meaningful feature spaces \cite{babenko2014neural,yu2017exploiting,gordo2016end}.
    
    In the following, this step is referred to as \textit{baseline retrieval} and will usually result in images that are all similar to the query, but with respect to different aspects of the query, so that they might not be similar compared to each other (cf.\ \cref{fig:aid-example}a). We assume that database items resembling the same aspect of the query are located in the same direction from the query in the feature space. Thus, we first represent all retrieved neighbors by their direction from the query:
    \begin{equation}
        \label{eq:direction-transform}
        \hat{X} \coloneqq
        \biggl\{\,
            \underbrace{\frac{x_i - q}{\| x_i - q \|}}_{\eqqcolon \hat{x}_i}
            \,\biggm\vert\, i = 1,\dots,m
        \,\biggr\} \;\;,
    \end{equation}
    where $\|\cdot\|$ denotes the Euclidean norm. Discarding the magnitude of feature vector differences and focusing on directions instead has proven to be beneficial for image retrieval, \eg, as so-called triangulation embedding \cite{jegou2014triangulation}.
    
    We then divide these directions $\hat{X}$ of the neighborhood into $k$ disjoint clusters (cf.\ \cref{fig:aid-example}b) using k-Means \cite{lloyd1982least}. For inspection by the user, each cluster is represented by a small set of $r$ images that belong to the cluster and are closest to the query $q$. This is in opposition to CLUE \cite{chen2005clue}, which represents each cluster by its medoid. However, this makes it difficult for the user to assess the relevance of the cluster, since the medoid has no direct relation to the query anymore.
    
    The proper number $k$ of clusters depends on the ambiguity of the query and also on the granularity of the search objective, because, for instance, more clusters are needed to distinguish between poodles and cocker spaniels than between dogs and other animals. Thus, there is no single adequate number of clusters for a certain query, but the same fixed value for $k$ is also likely to be less appropriate for some queries than for others. We hence use a heuristic found in literature for determining a query-dependent number of clusters based on the largest \textit{Eigengap} \cite{cai2004hierarchical}:
    \begin{enumerate}
        
        \item Construct an affinity matrix $A \in \mathbb{R}^{m \times m}$ with $A_{i,j} \coloneqq \exp\left(-\eta \cdot \| \hat{x}_i - \hat{x}_j \|^2\right)$.
        
        \item Compute the graph Laplacian $L = D - A$ with $D \coloneqq \diag(s_1, s_2, \dots, s_m)$, where $s_i \coloneqq \sum_{j=1}^{m} A_{i,j}$.
        
        \item Solve the generalized eigenvalue problem $L \mathbf{v} = \lambda D \mathbf{v}$ and sort the eigenvalues $\lambda_1, \lambda_2, \dots, \lambda_m$ in ascending order, \ie, $\lambda_1 \leq \lambda_2 \leq \dots \leq \lambda_m$.
        
        \item Set $k \coloneqq \argmax_{1 \leq i < m} ( \lambda_{i+1} - \lambda_i )$.
        
    \end{enumerate}
    
    \noindent
    This heuristic has originally been used in combination with spectral clustering, where the mentioned eigenvalue problem has to be solved as part of the clustering algorithm. Here, we use it just for determining the number of clusters and then apply k-Means as usual. The hyper-parameter $\eta$ can be used to control the granularity of the clusters: a smaller $\eta$ will result in fewer clusters on average, while large $\eta$ will lead to more clusters. In our experiments we set $\eta = \sqrt{d}$ and cap the number of clusters at a maximum of 10 to limit the effort imposed on the user.
    
    \subsection{Refinement of Results}
    \label{subsec:AID-refinement}
    
    Given a selection of $\ell$ relevant clusters represented by their centroids\footnote{Note that clustering has been performed on $\hat{X}$, so that the centroids represent (unnormalized) directions from the query as origin.} $C = \{c_1, \dots, c_\ell\}$, we re-rank all images in the database by adjusting their effective distance to the query, so that images in the same direction as the selected clusters are moved closer to the query, while images in the opposite direction are shifted away and images in the orthogonal direction keep their original scores. The images are then sorted according to this adjusted distance (cf.\ \cref{fig:aid-example}c).

    Let $x \in \mathcal{B}$ denote any image in the database, $\delta(x) \coloneqq \|x-q\|$ its Euclidean distance to the query (already computed during the initial retrieval) and $\sigma(x)$ the cosine similarity between the direction from $q$ to $x$ and from $q$ to the center of the relevant cluster closest to $x$, formally:
    \begin{equation}
        \label{eq:image-sim}
        \sigma(x) \coloneqq \max_{c_i \in C} \frac{c_i^\top (x-q)}{\|c_i\|\cdot\|x-q\|} \,.
    \end{equation}
    
    \noindent
    We define the adjusted distance score $\tilde{\delta}(x)$ of $x$ as
    \begin{equation}
        \label{eq:adjusted-distance}
        \tilde{\delta}(x) \coloneqq \delta(x) - \sign(\sigma(x)) \cdot |\sigma(x)|^\gamma \cdot \beta \,,
    \end{equation}
    where $\beta > 0$ is a constant that we set to $\beta \coloneqq \max_{x' \in \mathcal{B}} \delta(x')$ to ensure that even the most distant database item can be drawn to the query if it lies exactly in the selected direction. The hyper-parameter $\gamma \geq 0$ controls the influence of the user feedback: for $\gamma > 1$, only the distances of images matching the selected direction more exactly will be adjusted, while for $\gamma < 1$ peripheral images are affected as well. We consider $\gamma = 1.0$ a good default and use this in our experiments.
    
    Note that \cref{eq:adjusted-distance} allows for ``negative distances'', but this is not a problem, because we use the adjusted distance just for ranking and it is not a proper pair-wise metric anyway due to its query-dependence.

    \section{\uppercase{Experiments}}
    \label{sec:experiments}

    \begin{figure*}
        \includegraphics[width=\linewidth]{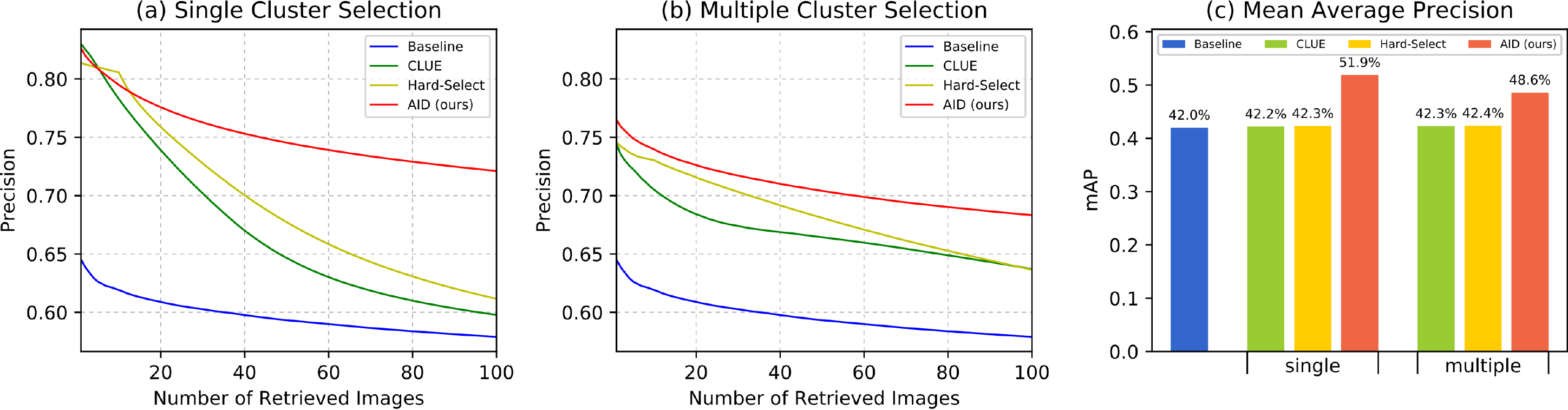}%
        \caption{%
            Performance of our AID approach compared with baseline retrieval, CLUE, and hard cluster selection on the same set of clusters as used by AID.%
            %Dotted lines and error bars indicate the 95\% confidence interval determined based on the standard deviation of all experimental runs.%
        }
        \label{fig:performance}
    \end{figure*}
    
    \subsection{Setup}
    \label{subsec:setup}
    
    \paragraph{Dataset}
    We evaluate the usefulness of our approach for image retrieval on the publicly available MIRFLICKR-25K dataset\footnote{\url{http://press.liacs.nl/mirflickr/}} \cite{huiskes08mirflickr}, which consists of 25,000 images collected from Flickr. All images have been annotated with a subset of 24 predefined topics by human annotators, where a topic is assigned to an image if it is at least somewhat relevant to it (``wide sense annotations''). A second set of annotations links topics to images only if the respective topic is saliently present in the image (``narrow sense annotations''), but these annotations are only available for 14 topics. Note that a single image may belong to multiple topics, which is in accordance with the ambiguity of query images.
    
    The median number of images assigned to such a ``narrow sense'' topic is 669, with the largest topic (``people'') containing 7,849 and the smallest one (``baby'') containing 116 images. Narrow sense topics are available for 12,681 images, which are on average assigned to 2 such topics, but at most to 5.
    
    We use all of those images to define 25,002 test-cases: Each image is issued as individual query for each of its assigned topics and the implied goal of the imaginary user is to find images belonging to the same topic. Due to the inherent ambiguity of a single query image, relevance feedback will be necessary in most cases to accomplish this task.
    
    \paragraph{Image Representations}
    Following the concept of \textit{Neural Codes} \cite{babenko2014neural}, we extract features for all images from a certain layer of a convolutional neural network. Specifically, we use the first fully-connected layer (\texttt{fc6}) of the VGG16 network \cite{simonyan2014vgg} and reduce the descriptors to 512 dimensions using PCA. We do explicitly not use features from the convolutional layers, although they have been shown to be superior for \textit{object} retrieval when aggregated properly \cite{babenko2015aggregating,zhi2016pmp}. This does, however, not hold for the quite different task of category retrieval, where the fully-connected layers---being closer to the class prediction layer and hence carrying more semantic information---provide better results \cite{yu2017exploiting}.
    
    \paragraph{Evaluation Metrics}
    Since the output of our image retrieval system is a ranked list of all images in the database, with the most relevant image at the top, we measure performance in terms of mean average precision (mAP) over all queries. Though this measure is adequate for capturing the quality of the entire ranking, it takes both precision and recall into account, whereas a typical user is seldom interested in retrieving all images belonging to a certain topic, but puts much more emphasis on the precision of the top results. Thus, we also report the precision of the top $\kappa$ results for $1 \le \kappa \le 100$.
    
    Because k-Means clustering is highly initializa\-tion-dependent, we have repeated all experiments 5 times and report the mean value of each performance metric. The standard deviation of the results was less than 0.1\% in all cases.
    
    \paragraph{Simulation of User Feedback}
    We investigate two different scenarios regarding user feedback: In the first scenario, the user must select exactly one of the proposed clusters and we simulate this by choosing the cluster whose set of preview images has the highest precision. In the second scenario, the user may choose multiple or even zero relevant clusters, which we simulate by selecting all clusters whose precision among the preview images is at least 50\%. If the user does not select any cluster, we do not perform any refinement, but return the baseline retrieval results.
    
    A set of $r = 10$ preview images is shown for each cluster, since ten images should be enough for assessing the quality of a cluster and we want to keep the number of images the user has to review as low as possible. Note that our feedback simulation does not have access to all images in a cluster for assessing its relevance, but to those preview images only, just like the end-user.
    
    \paragraph{The Competition}
    We do not only evaluate the gain in performance achieved by our AID method compared to the baseline retrieval, but also compare it with our own implementation\footnote{The source code of our implementation of AID and CLUE is available at \url{https://github.com/cvjena/aid}.} of CLUE \cite{chen2005clue}, which uses a different clustering strategy. Since CLUE does not propose any method to incorporate user feedback, we construct a refined ranking by simply moving the selected cluster(s) to the top of the list and then continuing with the clusters in the order determined by CLUE, which sorts clusters by their minimum distance to the query.
    
    For evaluation of the individual contributions of both our novel re-ranking method on the one hand and the different clustering scheme on the other hand, we also evaluate hard cluster selection (as used by CLUE) on the same set of clusters as determined by AID. In this scenario, the selected clusters are simply moved to the top of the ranking, leaving the order of images within clusters unmodified.
    
    %Both algorithms, CLUE and AID, use the $m = 200$ nearest neighbors of the query as input in the clustering stage.
    The number $m$ of nearest neighbors of the query used as input in the clustering stage should be large enough to include images from all possible meanings of the query, but larger $m$ also imply higher computational cost. We choose $m = 200$ as a trade-off for both, CLUE and AID.
    
    \subsection{Quantitative Results}
    \label{subsec:quantitative}
    
    The charts in \cref{fig:performance} show that our AID approach is able to improve the retrieval results significantly, given a minimum amount of user feedback. Re-ranking the entire database is of great benefit compared with simply restricting the final retrieval results to the selected cluster. The latter is done by CLUE and precludes it from retrieving relevant images not contained in the small set of initial results. Therefore, CLUE can only keep up with AID regarding the precision of the top 10 results, but cannot improve the precision of the following results or the mAP significantly.
    
    AID, in contrast, performs a global adjustment of the ranking, leading to a relative improvement of mAP over CLUE by 23\% and of P@100 by 21\%.
    
    The results for hard cluster selection on the same set of clusters as used by AID reveal that applying k-Means on $\hat{X}$ instead of $X$ (directions instead of absolute positions) is superior to the clustering scheme used by CLUE. However, there is still a significant gap of performance compared with AID, again underlining the importance of global re-ranking.
    
    Interestingly, though AID can handle the selection of multiple relevant clusters, it cannot take advantage from it, but multiple clusters even slightly reduce its performance (cf.\ \cref{fig:performance}b). This could not be remedied by varying $\gamma$ either and could be attributed to the fact that AID considers all selected clusters to be equally relevant, which may not be the case. If only the most relevant cluster is selected, in contrast, other relevant clusters will benefit from the adjusted distances as well according to their similarity to the selected one. This is supported by the fact that AID using a single relevant cluster is still superior to all methods allowing the selection of multiple clusters. Thus, we can indeed keep the required amount of user interaction at a minimum---asking the user to select a single relevant cluster only---while still providing considerably improved results.
    
    %In these experiments, we have determined the number of clusters individually for each query using the heuristic described in \cref{subsec:AID-clustering}. The average number of clusters proposed by that heuristic was 6.9 and the median was 7. Using a constant number of $k = 7$ clusters instead for all queries did not lead to significantly different results.
    %However, since no other fixed $k$ provided significantly better results, the employed heuristic seems appropriate for determining the number of clusters based on the data.
    
    \subsection{Qualitative Examples}
    \label{subsec:qualitative}
    
    \begin{figure*}[tb]
        \includegraphics[width=\linewidth]{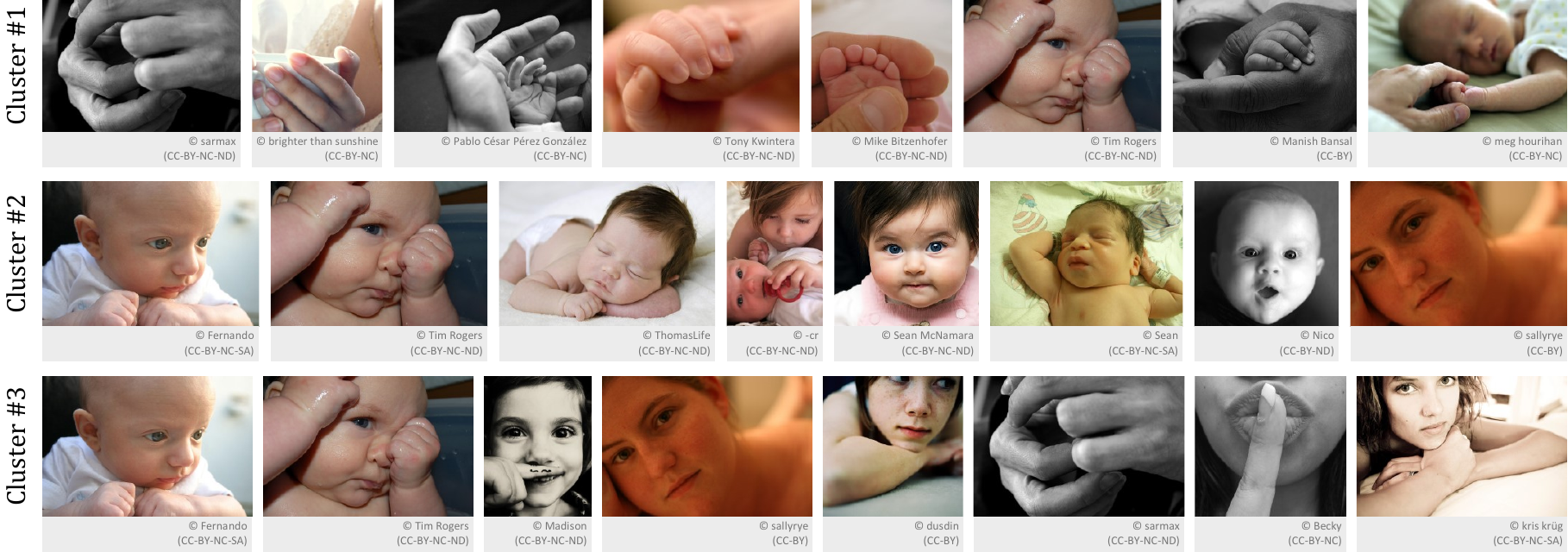}%
        \caption{Top refined results for the query from \cref{fig:ambiguity-examples}.}
        \label{fig:refined-examples-baby}
        %\vspace{10pt}
    \end{figure*}
    \begin{figure*}[tb]
        \includegraphics[width=\linewidth]{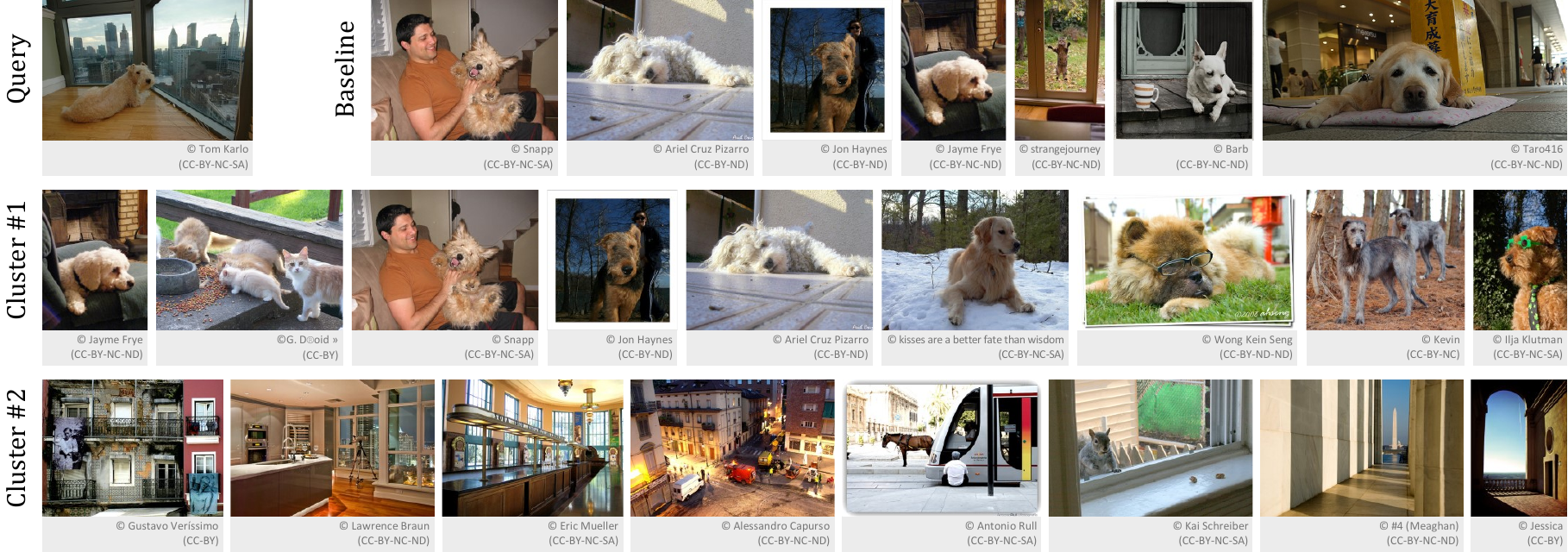}%
        \caption{Baseline and refined results for another query with two different meanings.}
        \label{fig:refined-examples-dog}
        %\vspace{20pt}
    \end{figure*}
    
    For an exemplary demonstration of our approach, we applied AID with a fixed number of $k = 3$ clusters to the query image from \cref{fig:ambiguity-examples}. The top 8 results from the refined ranking for each cluster are shown in \cref{fig:refined-examples-baby}.
    It can easily be observed that all clusters capture different aspects of the query: The first one corresponds to the topic ``hands'', the second to ``baby'', and the third to ``portrait''.
    
    Note that some images appear at the top of more than one refined ranking due to their high similarity to the query image. This is an advantage of AID compared with other approaches using hard cluster decisions, because the retrieved images might be just as ambiguous as queries and can belong to several topics. In this example, there is a natural overlap between the results for the topics ``baby'' and ``portrait'', but also between ``baby'' and ``hands'', since the hands of a baby are the prominent content of some images.
    
    A second example given in \cref{fig:refined-examples-dog} shows how AID distinguishes between two meanings of another query showing a dog in front of a city skyline. While the baseline ranking focuses on dogs, the results can as well be refined towards city and indoor scenes.

    \section{\uppercase{Conclusions}}
    \label{sec:conclusions}
    
    \noindent
    We have proposed a method for refining content-based image retrieval results with regard to the users' actual search objective based on a minimal amount of user feedback. Thanks to automatic disambiguation of the query image through clustering, the effort imposed on the user is reduced to the selection of a single relevant cluster. Using a novel global re-ranking method that adjusts the distance of all images in the database according to that feedback, we considerably improve on existing approaches that limit the retrieval results to the selected cluster.
    
    It remains as an open question, how feedback consisting of the selection of multiple clusters can be incorporated without falling behind the performance obtained from the selection of the single best cluster. Since some relevant clusters are more accurate than others, future work might investigate whether asking for a \textit{ranking} of relevant clusters can be beneficial.
    
    Furthermore, we are not entirely satisfied with the heuristic currently employed to determine the number of clusters, since it is inspired by spectral clustering, which we do not apply.
    %Moreover, a dynamic number of clusters per query should be able to outperform any fixed number of clusters for all queries.
    Since query images are often, but not always ambiguous, it would also be beneficial to detect when disambiguation is likely to be not necessary at all.

    \section*{Acknowledgments}
    \noindent
    This work was supported by the German Research Foundation as part of the
    priority programme ``Volunteered Geographic Information: Interpretation,
    Visualisation and Social Computing'' (SPP 1894, contract number DE 735/11-1).

    {\small
        \bibliographystyle{apalike}
        \bibliography{references}
    }

\end{document}